\documentclass{article} 
\usepackage{iclr2025_conference,times}
\usepackage{kotex}
\RequirePackage[format=plain,labelformat=simple,labelsep=period,font=small,compatibility=false]{caption}
\RequirePackage[font=footnotesize,skip=3pt,subrefformat=parens]{subcaption}

\definecolor{skyblue}{RGB}{100, 180, 240}
\definecolor{SkyEmerald}{RGB}{60,120,180}
\definecolor{dblue}{rgb}{0, 0, 0.6}
\definecolor{dgreen}{rgb}{0,0.4,0.2}
\definecolor{dred}{rgb}{0.6, 0, 0}
\definecolor{darkblue}{RGB}{0, 0, 139}
\usepackage[colorlinks,
            linkcolor=darkblue,
            anchorcolor=magenta,
            urlcolor=dgreen,
            citecolor=darkblue
           ]{hyperref}
\usepackage{url}
\usepackage{amsthm}
\usepackage{amsmath}
\usepackage{amssymb}
\usepackage{graphicx}
\usepackage{tikz-cd}
\usepackage{enumitem}
\usepackage{tcolorbox}
\usepackage{xcolor,colortbl}
\usepackage{bbm}
\usepackage[capitalise]{cleveref}
\usepackage{wrapfig,booktabs}

\usepackage{titletoc}
\usepackage{hyperref}
\definecolor{bluish}{rgb}{0, 0.4, 0.8}

\usepackage{amsmath,amsfonts,bm}

\def\eqref#1{equation~\ref{#1}}

\def\1{\bm{1}}

\DeclareMathAlphabet{\mathsfit}{\encodingdefault}{\sfdefault}{m}{sl}
\SetMathAlphabet{\mathsfit}{bold}{\encodingdefault}{\sfdefault}{bx}{n}

\usepackage{multirow}

\title{On the Internal Representations \\ of Graph Metanetworks}

\author{Taesun Yeom \& Jaeho Lee \\
Pohang University of Science and Technology
 (POSTECH)\\
\texttt{\{tsyeom,jaeho.lee\}@postech.ac.kr} \\
}

\iclrfinalcopy
\begin{document}

\maketitle

\begin{abstract}
Weight space learning is an emerging paradigm in the deep learning community. The primary goal of weight space learning is to extract informative features from a set of parameters using specially designed neural networks, often referred to as \emph{metanetworks}. However, it remains unclear how these metanetworks learn solely from parameters. To address this, we take the first step toward understanding \emph{representations} of metanetworks, specifically graph metanetworks (GMNs), which achieve state-of-the-art results in this field, using centered kernel alignment (CKA). Through various experiments, we reveal that GMNs and general neural networks (\textit{e.g.,} multi-layer perceptrons (MLPs) and convolutional neural networks (CNNs)) differ in terms of their representation space.
\end{abstract}

\section{Introduction}
\vspace{-0.5em}
\begin{wrapfigure}{R}{0.4\textwidth} 
\vspace{-1.4em}
    \centering
    \includegraphics[width=0.4\textwidth]{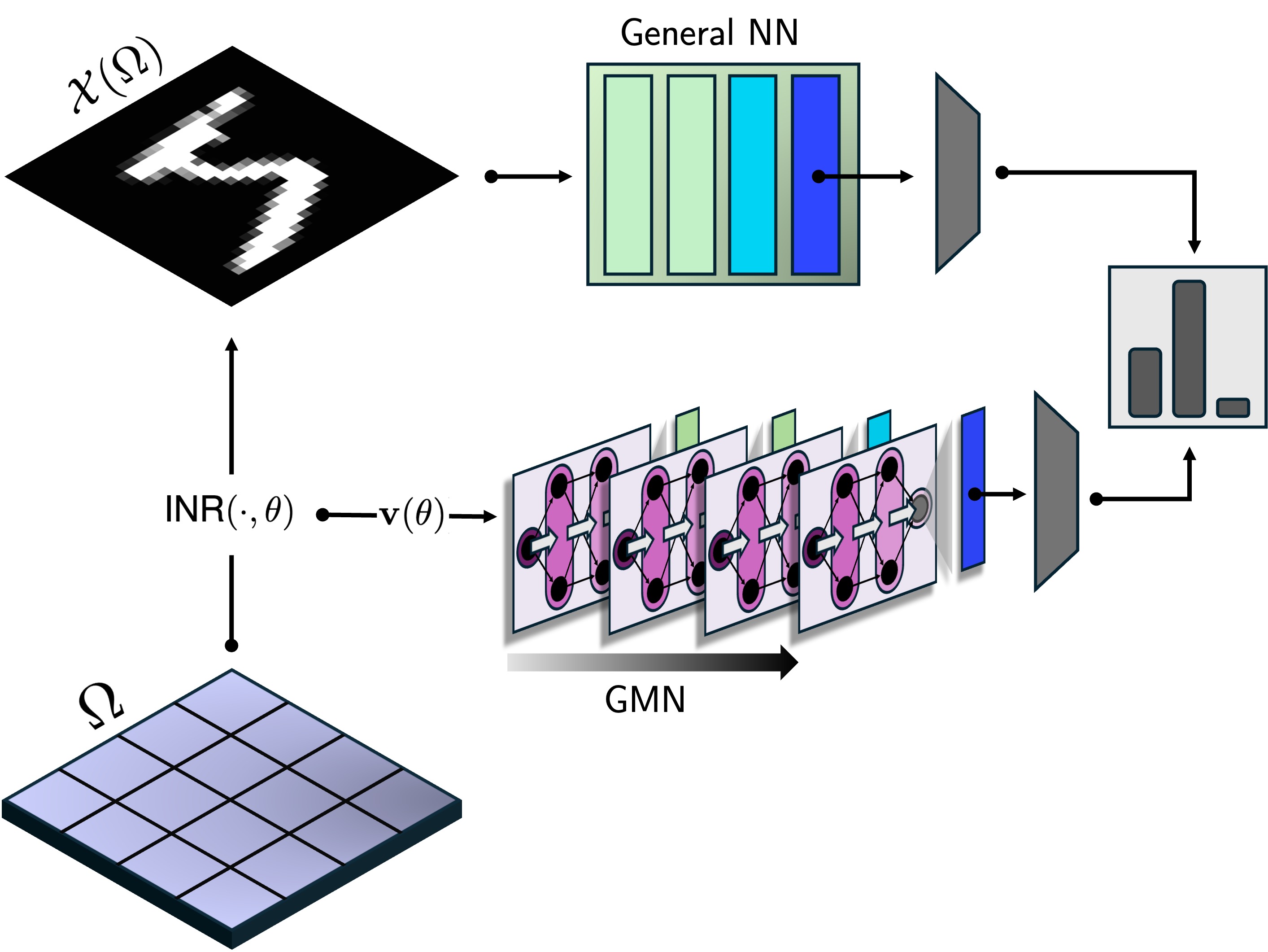}
    \caption{\textbf{GMN vs. General NN.}}
    \label{fig:main}
    \vspace{-1em}
\end{wrapfigure}

 The field of weight space learning has gained a lot of attention these days. Inspired by the concept of ``geometric deep learning'' \citep{bronstein2021geometric}, recent works have tended to focus on designing architectures that incorporate the symmetries underlying the parameter space of neural networks (NNs) (\textit{e.g.,} permutation, scaling, etc.) \citep{navon2023equivariant,zhou2023permutation}. These architectures are generally referred to as \emph{metanetworks}. Building on these seminal works, graph metanetworks (GMNs)—which process the parameters of neural networks as a graph structure—not only demonstrate superior performance across various tasks but also offer practical ease of use, as they can handle heterogeneous architectures and naturally accommodate permutation symmetry in the parameter space without the need for laborious equivariant layer design \citep{lim2024graph, kofinas2024graph}.

Despite recent advances in this field, state-of-the-art metanetworks (\textit{e.g.,} ScaleGMN \citep{kalogeropoulos2024scale}) have been observed to perform well on certain tasks (\textit{e.g.,} classifying MNIST INRs) but poorly on others (\textit{e.g.,} classifying CIFAR-10 INRs). This discrepancy naturally raises a  question: \emph{what do these types of networks really learn solely from the parameters of NNs?} Hence, in this work, we empirically investigate ``what GMNs learn'' through the lens of \emph{representations} (\textit{i.e.,} hidden features), which provide powerful insights for understanding NNs.

\textbf{Contributions.} Our main goal is to understand the representations in GMNs. To this end, we focus on an INR classification task for GMNs and, in turn, compare the representations they learn with those of MLPs and CNNs (which we refer to as ``general NNs'') performing standard image classification task (\cref{fig:main}). We begin by examining the effect of random initialization on representations in both cases and find that it leads to randomness in representations in GMNs, but not in general NNs. Next, we analyze the similarity of representations across different architectures and reveal that the representations of GMNs differ from those of general NNs through the cross-architecture CKA. Lastly, we find that this difference in representations indeed affects prediction tendencies, suggesting that weight space learning may play a complementary role in representation learning.

\section{Preliminaries}
\vspace{-0.5em}
In this section, we provide a brief overview of the similarity measure used in our work—centered kernel alignment (CKA)—along with a formal description of our setup. We note that the discussion of related work is provided in \cref{app:related_work}, and more details in \cref{app:details}.

\textbf{Centered kernel alignment (CKA).} CKA \citep{gretton2005measuring} is a nonparametric similarity measure that is invariant to orthogonal transformations and isotropic scaling. Let  \(X \in \mathbb{R}^{n \times d_X}\) and \(Y \in \mathbb{R}^{n \times d_Y}\) the representations of NNs, and \(\mathbf{X}\) and \(\mathbf{Y}\) the corresponding kernel matrices (\textit{e.g.,} \(\mathbf{X}_{ij} = K(X,X)_{ij}\)). In our work, we consider the kernel function \(K(\cdot,\cdot)\) to be linear (\textit{i.e.,} \(\mathbf{X} = XX^\top\)). Now, CKA is defined as
\begin{align}
    \text{CKA}(X,Y) = \frac{\text{H}(\mathbf{X},\mathbf{Y})}{\sqrt{\text{H}(\mathbf{X},\mathbf{X})\text{H}(\mathbf{Y},\mathbf{Y})}}, \quad \text{where} \quad \text{H}(\mathbf{X},\mathbf{Y})= \frac{1}{(n-1)^2}\text{tr}(\mathbf{X}C\mathbf{Y}C). \label{eq1}
\end{align}
In \cref{eq1}, $\text{H}(\cdot,\cdot)$ denotes the empirical estimator of Hilbert-Schmidt independence criterion (HSIC), and $C$ is a centering matrix (\textit{i.e.,} $C=I_n-\frac{1}{n}\bf{11}^\top$). CKA is widely used in deep learning, specifically as a similarity measure for representations across different initializations \citep{kornblith2019similarity}, architectures \citep{raghu2021vision}, and data modalities \citep{maniparambil2024vision}.

\textbf{GMNs vs. general NNs.} As mentioned in the previous section, we compare GMNs with ``general NNs'' (specifically, MLPs and CNNs in our case). The difference between GMNs and general NNs is straightforward: the latter directly take ``general data'' as input, whereas GMNs take only parameters as input. Formally, we consider a signal $\mathbf{x} \in \mathcal{X}(\Omega)$ that lies in a domain $\Omega$, and define `the function of an INR' as $\textsf{INR}(u,\theta): \Omega \to \mathcal{X}(\Omega)$, where $u$ is a set of discrete coordinates in $\Omega$, and $\theta$ is parameters of INR. While general NNs process $\mathbf{x}$ directly, GMNs process graph $\mathbf{v}(\theta) = (\mathcal{V}, \mathcal{E})$, referred to as \emph{neural graph}. Note that $\mathbf{v}(\cdot)$ is a function that converts $\theta$ into a (directed acyclic) graph.

\textbf{Our setup.} In this work, we consider a supervised classification task with three different datasets: MNIST, Fashion-MNIST, and CIFAR-10. Precisely, for each dataset containing $N$ data, we use \(D_{\text{NN}} = (\mathbf{x}_n, y_n)_{n=1}^{N}\) for general NNs and \(D_{\text{GMN}} = (\mathbf{v}(\theta_n), y_n)_{n=1}^{N}\) for GMNs. Here, $\mathbf{x_n}$ and $\mathbf{v}(\theta_n)$ serve as input for general NNs and GMNs each, and $y_n\in\mathbb{R}^k$ is the corresponding label vector with $k$ different classes. Note that $\theta_n$ represents the parameters of $\textsf{INR}_n(u_n, \theta_n)$, where the INR is trained to reconstruct $\mathbf{x}_n$. For GMNs, we perform an INR classification task where the GMN takes only $\mathbf{v}(\theta_n)$ as input to predict the label $y_n$. In contrast, general NNs learn to predict labels $y_n$ directly from the input image $\mathbf{x}_n$. Additionally, we implement INRs using `sinusoidal representation networks' (SIREN) architecture \citep{sitzmann2020implicit}, which employs a sine activation function. For GMNs, we adopt ScaleGMN \citep{kalogeropoulos2024scale} architecture. 

\section{On the internal representations of GMNs: an empirical study}
In this section, we begin our investigation by analyzing the impact of initialization on representations. Next, to gain deeper insights, we compare the representations of GMNs with those of general NNs. Lastly, we examine how representations influence predictions. Before proceeding, we introduce important details relevant to this section.
\begin{itemize}[leftmargin=*]
    \item For all experiments in this section, we use four hidden layers for general NNs and four MP steps for GMNs. Moreover, when comparing CKA between `different architectures,' we train general NNs until they achieve the highest test accuracy attained by GMNs for each dataset, since general neural networks always outperform GMNs when there are no constraints in the classification task.
    \item To calculate CKA, we use the activation matrix obtained from the test dataset at the $k$-th hidden layer in general NNs, and the features of the output node(s) (\textit{e.g.,} gray node in \cref{fig:main}) after $k$ MP steps in GMNs, both of which we set to have the same dimensions for per task. Here, we set $k \in \{\textcolor{skyblue}{3}, \textcolor{blue}{4}\}$ due to the learning strategy of GMNs\footnote{Since we consider a three-layer INR, we need at least three forward MP steps to propagate information from the input node to the output nodes. That is, for GMNs to ``learn representations of the whole graph,'' they require at least three update steps.}.

\end{itemize}

\subsection{On the impact of random initialization} \label{sec:random}
\vspace{-0.2em}
It is known that random initialization plays a crucial role in representations of NNs from various perspectives (see \cref{app:related_work}). Building on this, we first study how random initialization in NNs affects their representations. We trained ten NNs with different initializations for each architecture and paired them within the same architecture, forming five pairs. For every training epoch, we compute the CKA for each pair using the test dataset and take the average per architecture, which we denote as $\overline{\text{CKA}}(k,\textbf{a})$ for brevity.

\begin{figure}[t]
    \centering
    \begin{subfigure}{0.32\textwidth}
        \centering
        \includegraphics[width=\textwidth]{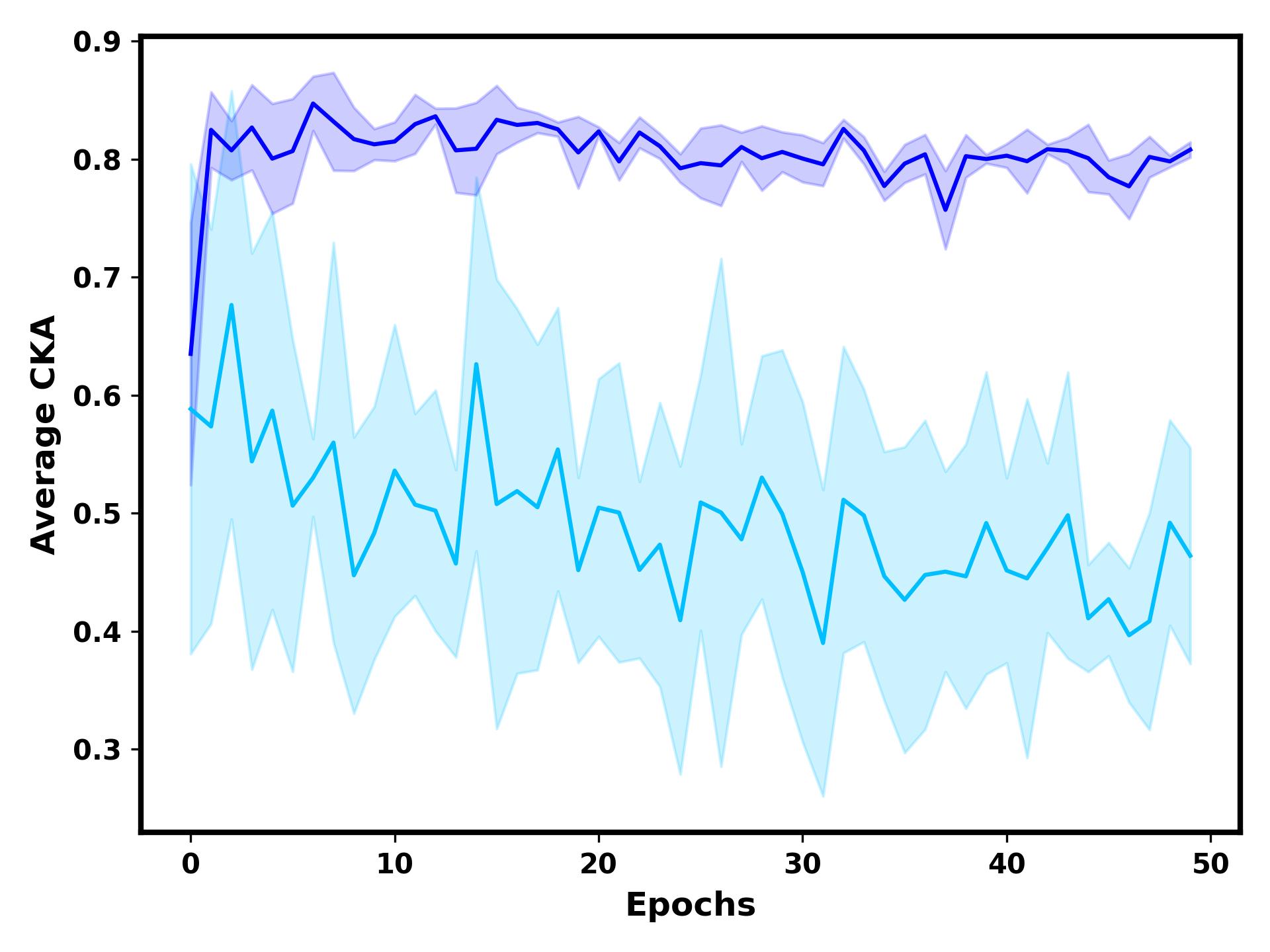}
    \caption{GMN (MNIST INRs)}
    \label{same_net:GMN}
    \end{subfigure}
    \centering
    \begin{subfigure}{0.32\textwidth}
        \centering
        \includegraphics[width=\textwidth]{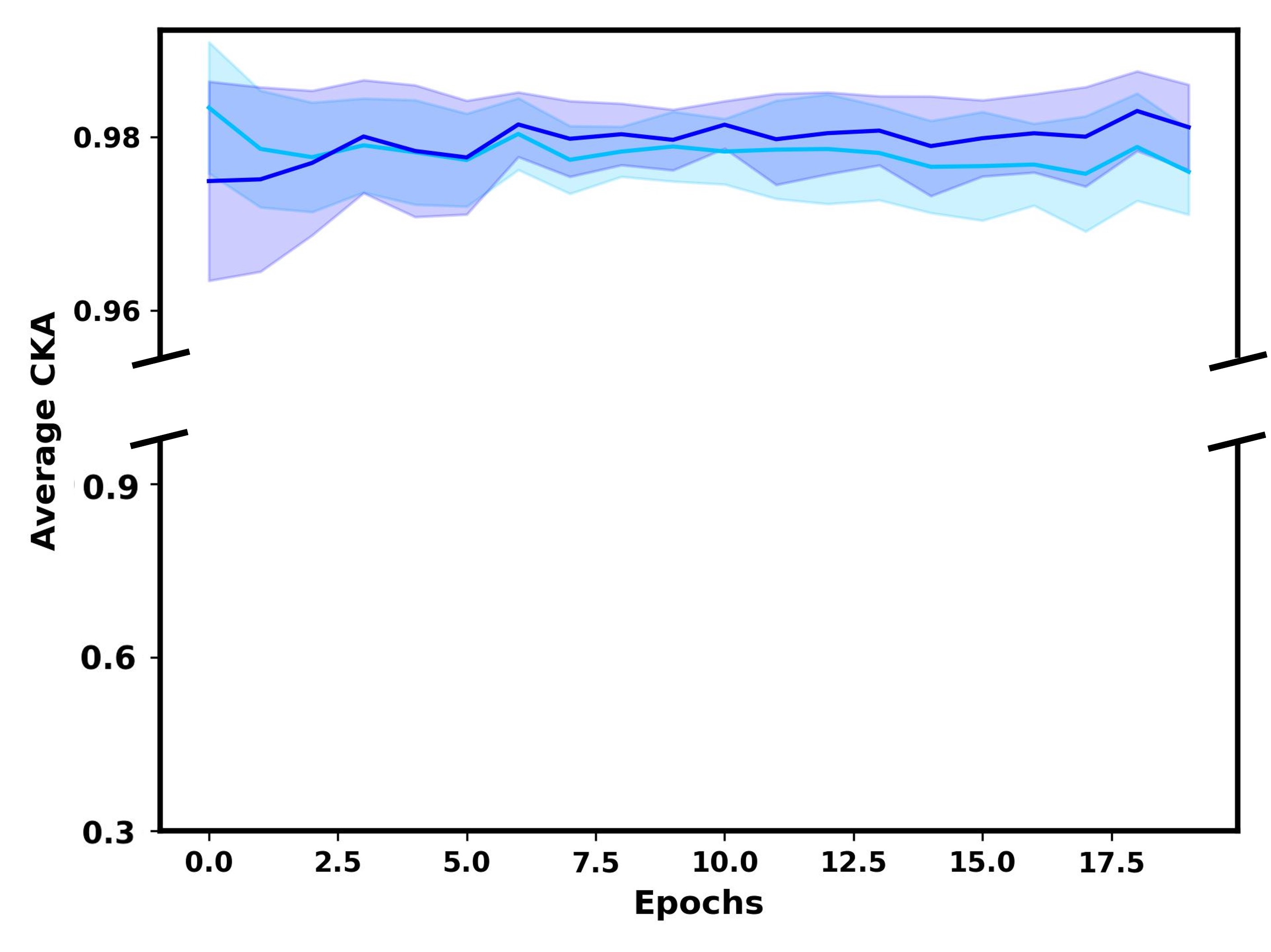}
    \caption{MLP (MNIST images)}
    \label{same_net:MLP}
    \end{subfigure}
    \centering
    \begin{subfigure}{0.32\textwidth}
        \centering
        \includegraphics[width=\textwidth]{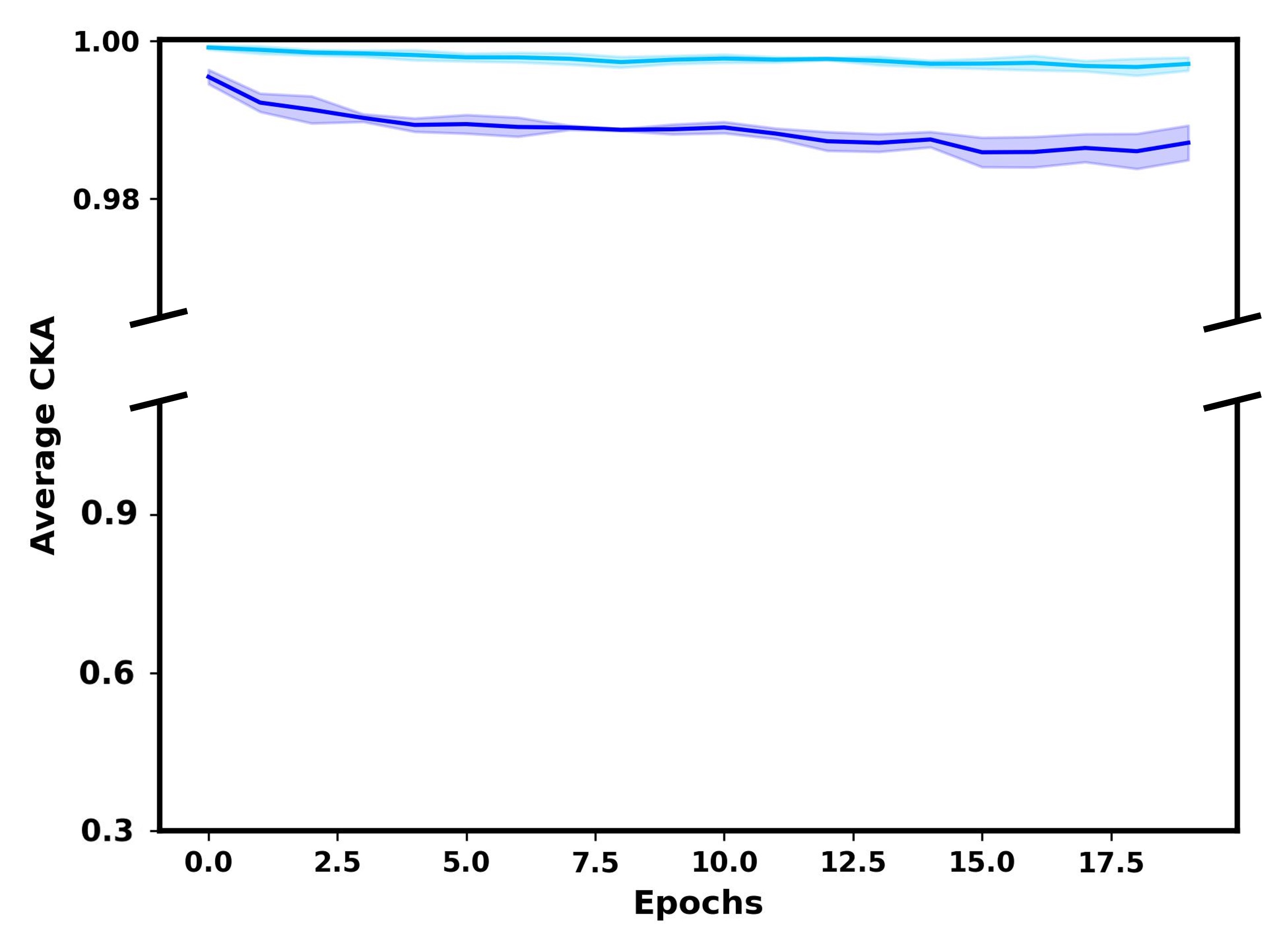}
    \caption{CNN (MNIST images)}
    \label{same_net:CNN}
    \end{subfigure}
    \vspace{-0.5em}
\caption{\textbf{Averaged CKA over five pairs of random GMNs, MLPs, and CNNs in MNIST classification.} 
The \textcolor{skyblue}{skyblue} plot represents the averaged $\overline{\text{CKA}}(\textcolor{skyblue}{3},\textbf{a})$, while the \textcolor{blue}{blue} plot represents the averaged $\overline{\text{CKA}}(\textcolor{blue}{4},\textbf{a})$ for each architecture.}
    \label{fig:same_net}
\end{figure}

In \cref{fig:same_net}, we plot the dynamics of $\overline{\text{CKA}}$ throughout training. Observing the results for MLP and CNN cases, we see that random initialization does not significantly impact the representation similarity of general NNs, as indicated by consistently high CKA values (\textit{i.e.,} between 0.9 and 1), regardless of the layer index. However, in the case of GMNs (\cref{same_net:GMN}), $\overline{\text{CKA}}(\textcolor{skyblue}{3},\text{GMN})$ is significantly lower during training compared to general NNs. This may be due to the presence of projection layers in GMNs, which project node and edge features (which are originally scalar values) to a higher dimension using multi-layer linear networks. The intriguing part is that the representations do not converge (or at least do not increase) with each other throughout training. This  indicates that ``different initializations lead to different hidden representations in GMNs throughout training,'' which may substantially affect the representations of the last step of MP. For instance, $\overline{\text{CKA}}(\textcolor{blue}{4},\text{GMN})$ are considerably lower than those of MLPs and CNNs (\textit{i.e.,} almost 1) throughout training. Note that lower CKA does not reflect higher variance in classification accuracy (\cref{tab:det} in the Appendix).

{\bf\textit{Takeaways.}} Unlike general NNs, the effect of random initialization in GMNs is significant: it reduces the similarity of representations. In other words, random GMNs captures different representations depending on initialization.

\subsection{GMNs and general NNs learn different representations}

\begin{wraptable}{t}{0.5\textwidth} 
\centering
\resizebox{0.5\textwidth}{!}{%
\begin{tabular}{@{}l c c c c@{}}
\toprule
\multirow{2}{*}{} & \multirow{2}{*}{} & \multicolumn{3}{c}{Dataset} \\ \cmidrule(l){3-5} 
                &                   & MNIST & Fashion-MNIST & CIFAR-10 \\ \midrule
\multirow{3}{*}{CKA} 
                & GMN $\leftrightarrow$ MLP  & 0.42 $\pm$ 0.04 & 0.42 $\pm$ 0.03 & 0.38 $\pm$ 0.06 \\
                & GMN $\leftrightarrow$ CNN  & 0.43 $\pm$ 0.03 & 0.49 $\pm$ 0.02 & 0.43 $\pm$ 0.05 \\
                & MLP $\leftrightarrow$ CNN  & 0.94 $\pm$ 0.01 & 0.95 $\pm$ 0.01 & 0.94 $\pm$ 0.01 \\ \midrule
\multirow{6}{*}{NHD} 
                & Random GMNs    & 0.041 $\pm$ 0.002 & 0.137 $\pm$ 0.006 & 0.229 $\pm$ 0.007 \\
                & Random MLPs    & 0.022 $\pm$ 0.001 & 0.045 $\pm$ 0.007 & 0.062 $\pm$ 0.009 \\
                & Random CNNs    & 0.023 $\pm$ 0.002 & 0.036 $\pm$ 0.002 & 0.062 $\pm$ 0.004 \\ \cmidrule(l){2-5} 
                & GMN $\leftrightarrow$ MLP  & 0.050 $\pm$ 0.002 & 0.192 $\pm$ 0.007 & 0.349 $\pm$ 0.013 \\
                & GMN $\leftrightarrow$ CNN  & 0.058 $\pm$ 0.002 & 0.175 $\pm$ 0.006 & 0.350 $\pm$ 0.013 \\
                & MLP $\leftrightarrow$ CNN  & 0.033 $\pm$ 0.001 & 0.093 $\pm$ 0.004 & 0.145 $\pm$ 0.009 \\ \bottomrule
\end{tabular}%
}
\vspace{-0.5em}
\caption{\textbf{Averaged CKA between different architectures and NHD from various cases.}}
\label{tab:main_tab}
\end{wraptable}
\vspace{-0.4em}
Next, we conduct experiments to compare representations across architectures. Since the learning curve of test accuracy varies depending on both architecture and dataset, we compare representations only at the final epoch for a fair comparison. Additionally, we analyze the representations just before the linear classifier (\textit{i.e.,} $k=\textcolor{blue}{4}$). Similar to the previous subsection, we compare five pairs within each group. However, in this case, we consider three different groups: GMN vs. MLP, GMN vs. CNN, and MLP vs. CNN. Here, we denote $\overline{\text{CKA}}(k,\mathbf{a} \leftrightarrow \mathbf{b})$ as the averaged CKA between two types of networks, $\mathbf{a}$ and $\mathbf{b}$.

In \cref{tab:main_tab} (CKA), we report $\overline{\text{CKA}}(\textcolor{blue}{4},\mathbf{a} \leftrightarrow \mathbf{b})$ along with its standard deviation, computed within each group. Interestingly, despite the three different network types achieving nearly the same test accuracy with the same classifier structure, the CKA value heavily depends on the combination of architectures. For example, when comparing representations in general NNs (\textit{i.e.,} MLPs vs. CNNs), the CKA value remains consistently high across various datasets. However, when comparing architectures from different groups (\textit{i.e.,} GMNs vs. \{MLPs, CNNs\}), the CKA value is considerably lower than in the former case.

It is notable that, despite MLPs and CNNs being known to have different inductive biases, their final representations are still somewhat aligned. On the contrary, GMNs behave differently from general NNs, even though their test accuracies are almost identical to those of general NNs (see \cref{tab:det}). These results suggest that the representations of GMNs and general NNs differ, \textit{i.e.,} the representations learned from parameters and images are different.

{\bf\textit{Takeaways.}} By examining CKAs between GMNs and general NNs, we can see that GMNs learn representations differently from general NNs, whereas CNNs and MLPs learn in a similar manner.

\subsection{On the prediction tendency of GMNs} \label{sec:tendency}
\vspace{-0.4em}
A natural question arises here: if GMNs and general NNs learn different representations, how do they achieve almost same test accuracies? While it is possible for different feature spaces to yield the same classification results, it is important to examine this from the perspective of predictions—specifically, whether the predictions made by GMNs are genuinely similar or different to those of general NNs.

To this end, we evaluate the normalized Hamming distance (NHD) \citep{hamming1950error}, which quantifies the difference between two sequences that consist of binary elements: each element in the sequence is defined as `1' (if the prediction is correct) or `0' (if it is not)\footnote{NHD is computed as $d_H(p, q) = \frac{1}{N} \sum_{i=1}^{N} \mathbbm{1}_{q_i \neq p_i}$, where $p=(a_1, \cdots, a_N)$ and $q=(b_1, \cdots, b_N)$ are the two sequences of networks' predictions being compared, and $N$ is the length of the test dataset.}. We report the averaged NHDs in \cref{tab:main_tab}, which are obtained by averaging the NHDs over five pairs for each case. First, we analyze the impact of random initialization on predictions (denoted as ``Random networks''). In this case, GMNs exhibit a considerably higher NHD than general NNs, meaning that the predictions of GMNs are more variable. Furthermore, we evaluate the NHD across different architectures (denoted as ``$\mathbf{a}\leftrightarrow \mathbf{b}$''). It can be seen that the NHDs between MLPs and CNNs are low, whereas those between GMNs and general NNs are relatively high. This result indicates that the predictions made by GMNs and general NNs indeed differ, while MLPs and CNNs yield nearly identical predictions.

For an in-depth analysis, we investigate whether there are instances in which GMNs make correct predictions while general NNs fail in MNIST classification. We examine this in both our original setting—where general NNs achieve nearly the same test accuracy as GMN—and an additional setting in which general NNs are trained longer to reach better test accuracy (details are in \cref{tab:det}). Surprisingly, in both cases, there exist instances that which GMN consistently makes correct predictions, whereas MLP and CNN fail across all cases (\textit{e.g.,} \cref{fig:incorrect}), even if the lower test accuracy. This phenomenon implies that the representations of GMNs learned from ``parameters'' may complement the representations of general NNs learned from ``visual signals.''

{\bf\textit{Takeaways.}}  The different internal representations of GMNs and general NNs indeed impact predictions. Moreover, in some cases, there are samples that only GMNs can correctly classify.

\section{Conclusions}
\vspace{-0.4em}
In this work, we empirically investigate the representations of GMNs using CKA, comparing them to general NNs, specifically MLPs and CNNs. We find that random initialization has a significantly greater impact on the representations of GMNs than on those of general NNs. Furthermore, while general NNs tend to learn similar features, GMNs exhibit distinct representations, suggesting that they tend to capture different types of information depending on initialization. Finally, these differences in representation ultimately lead to differences in predictions.

\textbf{Limitations and future directions.} We focus on only one task of weight space learning (\textit{i.e.,} INR classification), where the task does not reflect equivariance. Another limitation lies in our selection of general NNs, which we exclude widely used models, such as vision transformer \citep{dosovitskiy2021an}. However, we strongly believe that our work will inspire promising future directions for improving the performance of metanetworks, as well as other applications of them.



\newpage
\bibliography{./iclr2025_conference}
\bibliographystyle{iclr2025_conference}

\newpage
\appendix
{\Large \scshape \raggedright
Appendix\par
}
\section{Related work} \label{app:related_work}
\textbf{Weight space learning and metanetworks.} The study of weight space learning begins with the observation that statistics of neural network parameters contain information (\textit{e.g.,} generalization properties of NN) \citep{unterthiner2020predicting}. The next approaches involve building neural methods for deep learning on weights for various downstream tasks, where these type of architectures are referred to as ``metanetworks'' or ``neural functionals''. The major building blocks of metanetworks (MNs) involve designing architectures that account for the symmetries of weight space (\textit{e.g.,} permutation, scaling, etc.). The early works in MNs consider applying permutation-equivariant layers \citep{navon2023equivariant, zhou2023permutation}, based on blueprint of ``geometric deep learning'' \citep{bronstein2021geometric}. But these approaches are often not ``versatile'' since they cannot handle heterogeneous architectures in a dataset. More recent approaches focus on better representations of parameters. \citet{lim2024graph} and \citet{kofinas2024graph} suggest representing parameters as a ``graph'' and learning representations using graph neural networks, which is called as graph metanetworks (GMNs). \citet{kalogeropoulos2024scale} improve GMNs by considering additional symmetries inherent in neural network weights, specifically scaling (or sign) symmetry. While these GMNs outperform prior works on various tasks, a major concern remains unaddressed: \emph{GMNs still perform poorly compared to traditional data-based neural networks and why}. To fill this gap, it is crucial to understand GMNs and the difference between data-based and parameter-based learning. In line of this, we take a first step toward understanding GMNs, specifically from a representation perspective.

\textbf{Implicit neural representations as data.}
Implicit neural representations (INRs) are a family of neural networks that map spatiotemporal coordinates to signal values, \textit{e.g.,} image INRs learn a mapping from coordinates in a 2D Euclidean domain to RGB values. Since an INR represents a single datum, some recent works consider each INR as a datum itself—more precisely, its parameters as a datum. The paradigm of ``INRs as data'' offer the possibility of storing continuous signal in a efficient manner \citep{dupont2022generative,dupont2022data}, which is totally different from traditional data representations. \citet{luigi2023deep} uses an auto-encoding scheme to learn a latent space of INR functions solely from parameters, which are input as flattened vectors. \citet{papa2024train} studies the trade-off between the ``reconstruction quality of INRs'' and the ``learnability from INR parameters.'' \citet{yeom2025fast} analyzes the effect of the initialization scheme (\textit{i.e.,} variance of initialization distribution) of INRs on learning in weight space. Despite this line of work, it has been observed that state-of-the-art INR-processing networks (\textit{i.e.,} \citet{kalogeropoulos2024scale}) perform poorly on some tasks. For instance, CIFAR-10 INR classification achieves only about 37\%, compared to CIFAR-10 image classification on traditional neural networks, which reaches nearly 99\%, while the reasons remain unknown. In this work, we study the phenomena arising in INR classification (using GMNs) versus image classification (using MLPs and CNNs), specifically in terms of their internal representations.

\textbf{Internal representations of neural networks.}
As part of efforts to understand NNs, interest in their internal representations has persisted over time. \citet{kornblith2019similarity} study the activation matrices of CNNs and reveal that differently initialized CNNs learn similar features. \citet{raghu2021vision} comprehensively evaluate the representations between CNNs (\textit{e.g.,} ResNets \citep{he2016deep}) and Vision Transformers (ViTs) \citep{dosovitskiy2021an}, finding correspondences between blocks or layers. \citet{maniparambil2024vision} reveal that text encoders and image encoders share representations, as indicated by higher CKA values, a similar observation also noted in \citet{huhposition}. In the field of metanetworks, \citet{shamsianimproved} study data augmentation in weight space to improve generalization of metanetworks.

\newpage
\section{Experimental details} \label{app:details}
\subsection{Dataset details} \label{app:dataset_details}
We consider three different datasets for experiments, details are as follows.
\begin{itemize}[leftmargin=*]
    \item \textbf{MNIST:} We use MNIST image dataset for general NNs, and MNIST INR dataset from \citet{navon2023equivariant}. Each dataset contains 55k training images/INRs, 5k validation images/INRs, and 10k test images/INRs. 
    \item \textbf{Fashion-MNIST:} We use Fashion-MNIST image dataset for general NNs, and Fashion-MNIST INR dataset also from \citet{navon2023equivariant}. Each dataset contains 55k training images/INRs, 5k validation images/INRs, and 10k test images/INRs.  
    \item \textbf{CIFAR-10:} We use CIFAR-10 image dataset for general NNs, and CIFAR-10 INR dataset from \citet{zhou2023permutation}. Each dataset contains 45k training images/INRs, 5k validation images/INRs, and 10k test images/INRs. 
\end{itemize}
For all experiments, we report ``test accuracy,'' where all test datasets are class-balanced. For INRs, we use three-layer SIREN \citep{sitzmann2020implicit}, with 32 as a hidden dimension.

\subsection{Network details} \label{app:net_details}
\begin{itemize}[leftmargin=*]
    \item \textbf{GMN:} We adopt ScaleGMN \citep{kalogeropoulos2024scale} for a architecture of GMN. Specifically, we use default setup of ScaleGMN without bidirectional message passing. Precisely, each node (\textit{i.e.,} bias) and directed edge (\textit{i.e.,} weight) of the neural graph are projected into a higher dimensional feature space via linear networks and trained using a message passing algorithm designed to account for the sign symmetries in SIREN INR (Section 5 in \citet{kalogeropoulos2024scale}). Aggregation occurs at the output nodes of neural graphs after four steps of message passing. Here, we refer to ``the updated features of output node(s) after $k$ steps'' as ``$k$-th hidden representations,'' which share a similar concept with activation matrices in general NNs.
    \item \textbf{MLP:} We use simple four-hidden layer MLP architecture, with 128 (for MNIST and Fashion-MNIST) or 192 (for CIFAR-10) hidden dimension with ReLU activation functions.
    \item \textbf{CNN:} In this case, we also use a simple four-hidden layer CNN architecture. Convolutional layers consist with (1) 128 (for MNIST and Fashion-MNIST) or 192 (for CIFAR-10) $7\times7$ filters with stride 3 and (2,3,4) 128 (for MNIST and Fashion-MNIST) or 192 (for CIFAR-10) $2\times2$ filters with stride 2. Since the activation matrices of CNNs have multiple channels, we apply average pooling if the number of channels is greater than one.
\end{itemize}

\subsection{Additional details} \label{app:add_details}
We train each networks for ten different seeds with AdamW optimizer. Additionally, for fair comparison between representations, we trained general neural networks until they reached the best test accuracy of GMNs for each task. We report test accuracies, training epochs, and learning rates in \cref{tab:det}.
\begin{table}[]
\centering
\resizebox{0.85\textwidth}{!}{ 
\begin{tabular}{@{}lccc@{}}
\toprule
                               & GMN (INR input)      & MLP (image input)                                 & CNN (image input)                              \\ \midrule
\multirow{2}{*}{MNIST}         & 95.92 $\pm$ 0.31 & 96.13 $\pm$ 0.24 $\rightarrow$ \textcolor{blue}{98.74 $\pm$ 0.11} & 96.06 $\pm$ 0.27 $\rightarrow$ \textcolor{blue}{98.31 $\pm$ 0.11}             \\
                               & [50, $1e-03$]  & [10, $1e-04$] $\rightarrow$ \textcolor{blue}{[20, $1e-04$]}    & [5, $5e-05$] $\rightarrow$ \textcolor{blue}{[20, $5e-05$]} \\ \midrule
\multirow{2}{*}{Fashion-MNIST} & 77.83 $\pm$ 0.63 & 78.19 $\pm$ 0.42                              & 79.60 $\pm$ 0.41                           \\
                               & [30, $1e-03$]  & [20, $1e-05$]                               & [20, $1e-05$]                            \\ \midrule
\multirow{2}{*}{CIFAR-10}      & 35.73 $\pm$ 1.23 & 36.91 $\pm$ 0.45                              & 38.02 $\pm$ 0.36                           \\
                               & [100, $1e-03$] & [25, $1e-05$]                               & [10, $1e-05$]                            \\ \bottomrule
\end{tabular}
}
\caption{\textbf{Test accuracies ($\%$) across datasets and tasks \& hyperparameters.} 
Values in [square brackets] represent [training epochs, learning rate], respectively. 
We additionally report details of more intense training in \cref{sec:random} and \cref{sec:tendency} as \textcolor{blue}{blue}.}
\label{tab:det}
\end{table}

\newpage
\section{Additional results} \label{app:add_result}

\begin{figure}[h]
    \centering
    \begin{subfigure}{0.32\textwidth}
        \centering
        \includegraphics[width=\textwidth]{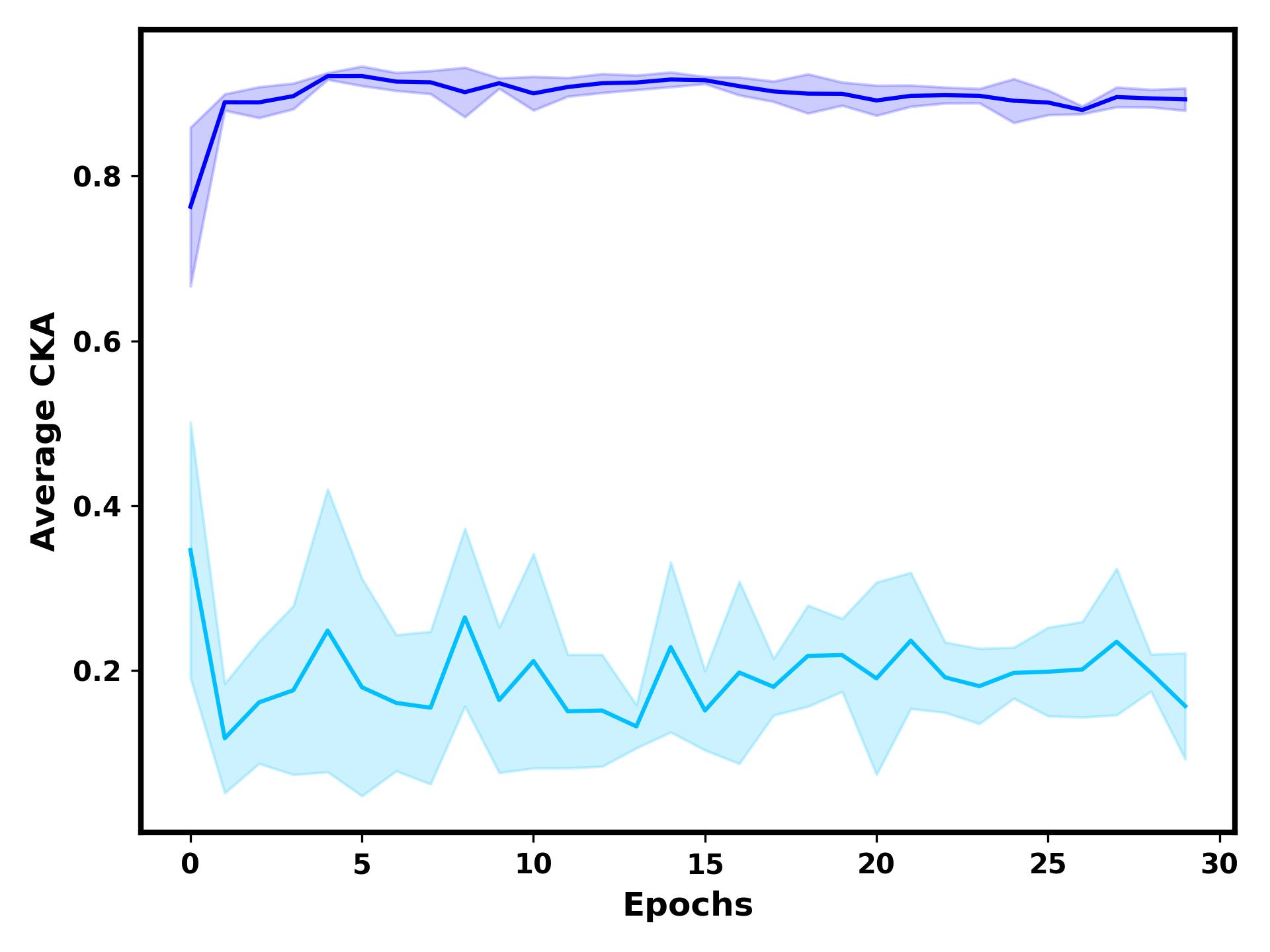}
        \caption{GMN (Fashion-MNIST INRs)}
    \end{subfigure}
    \begin{subfigure}{0.32\textwidth}
        \centering
        \includegraphics[width=\textwidth]{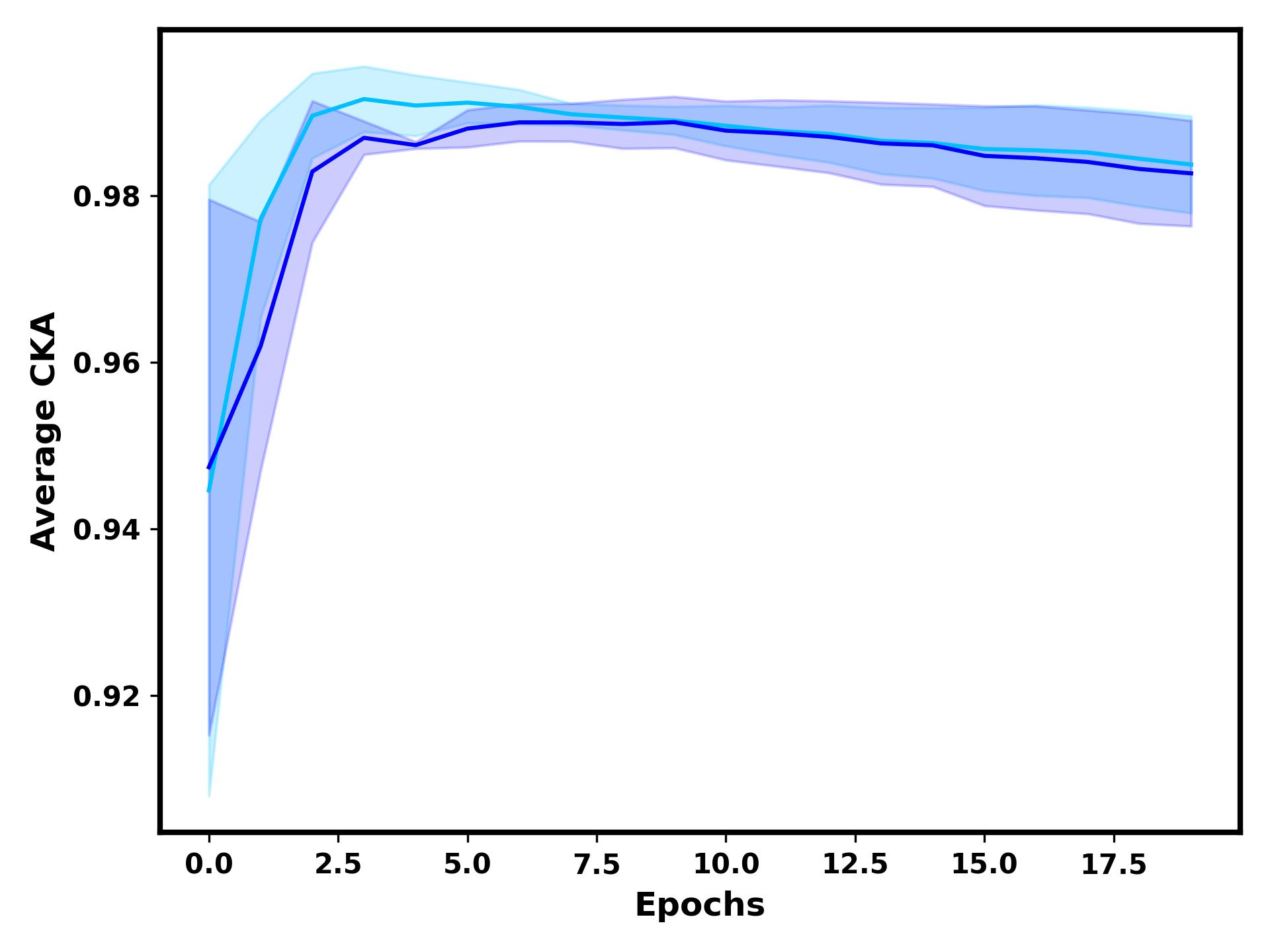}
        \caption{MLP (Fashion-MNIST images)}
    \end{subfigure}
    \begin{subfigure}{0.32\textwidth}
        \centering
        \includegraphics[width=\textwidth]{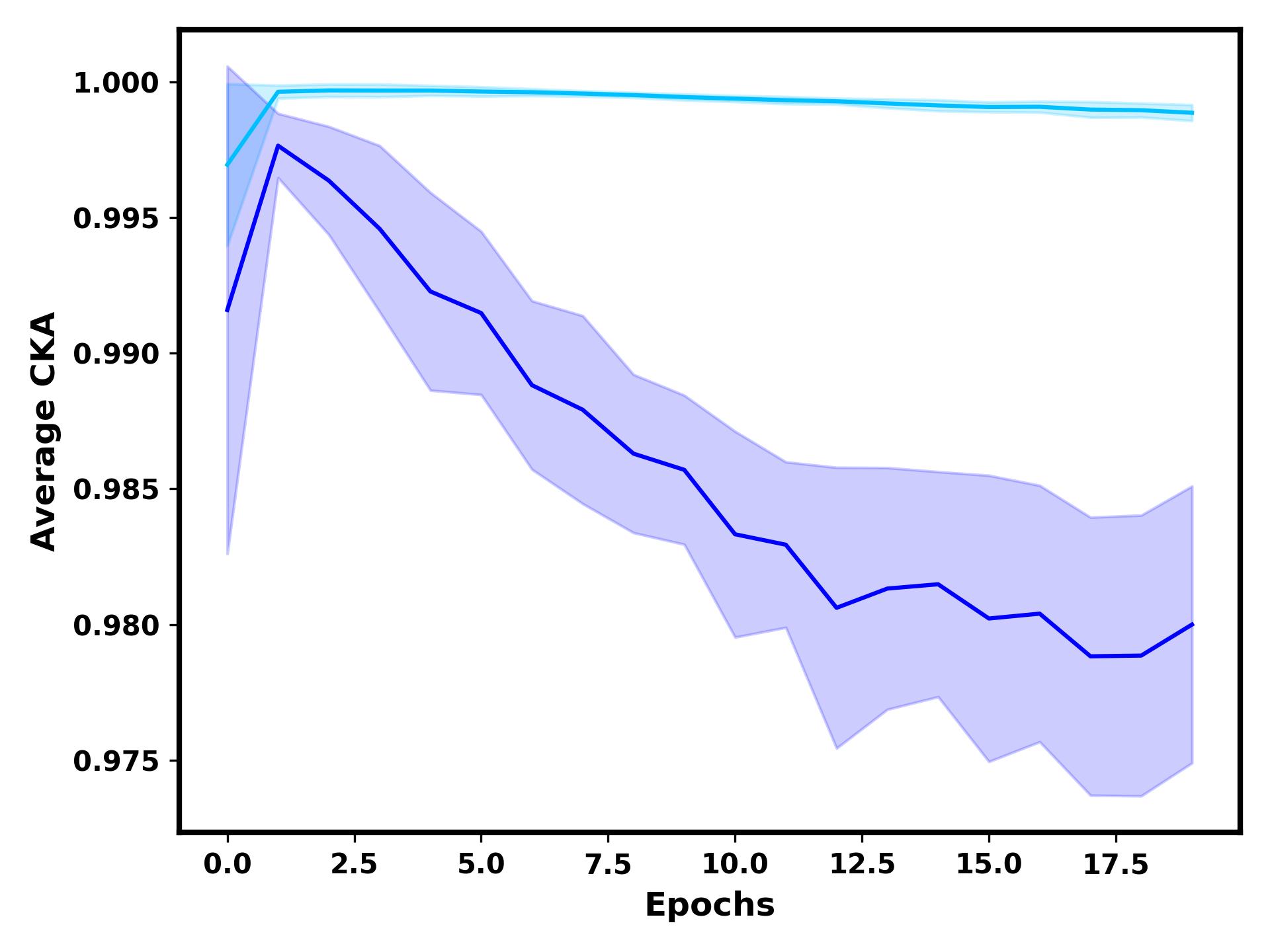}
        \caption{CNN (Fashion-MNIST images)}
    \end{subfigure}
    
    \begin{subfigure}{0.32\textwidth}
        \centering
        \includegraphics[width=\textwidth]{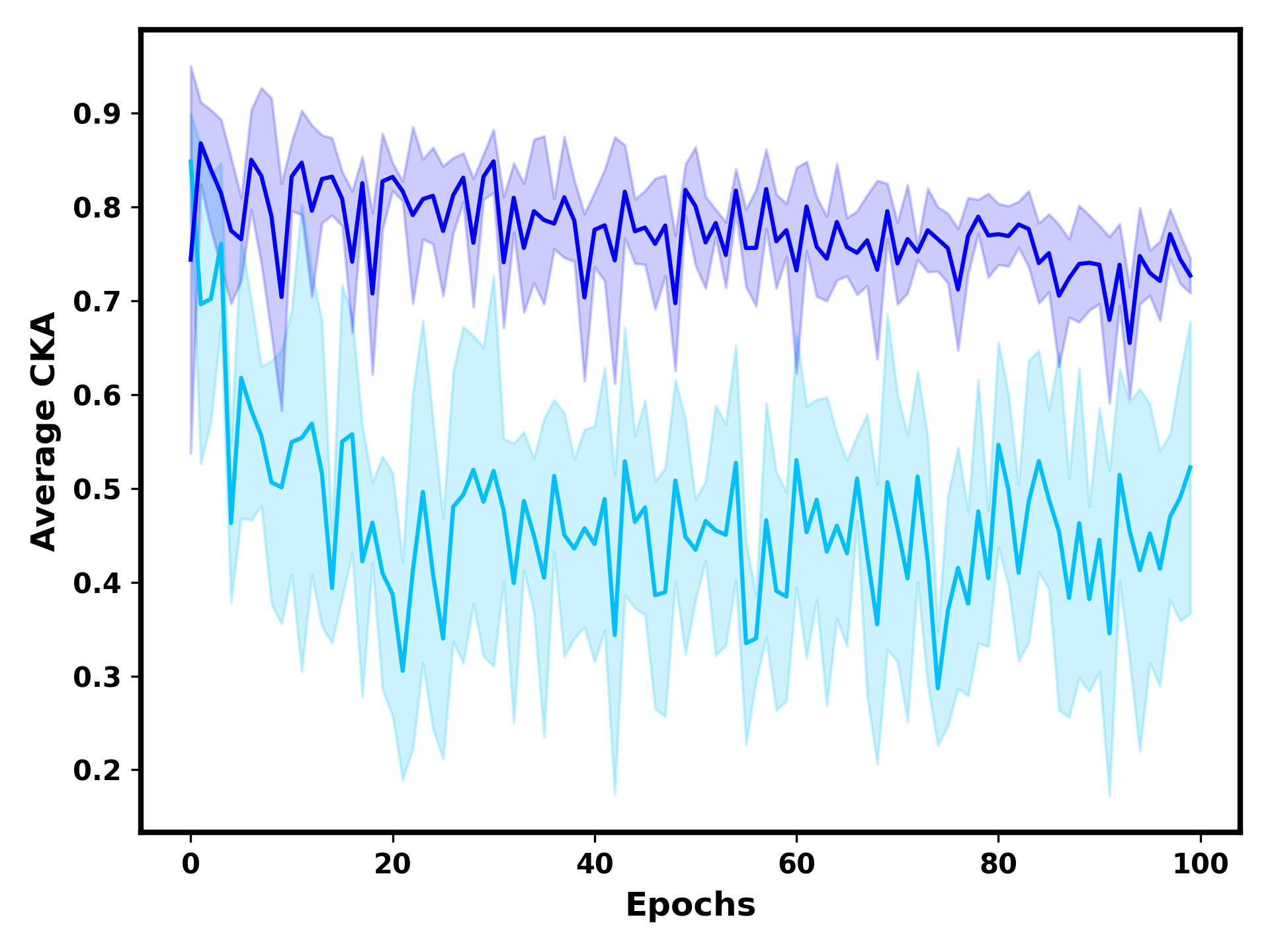}
        \caption{GMN (CIFAR-10 INRs)}
    \end{subfigure}
    \begin{subfigure}{0.32\textwidth}
        \centering
        \includegraphics[width=\textwidth]{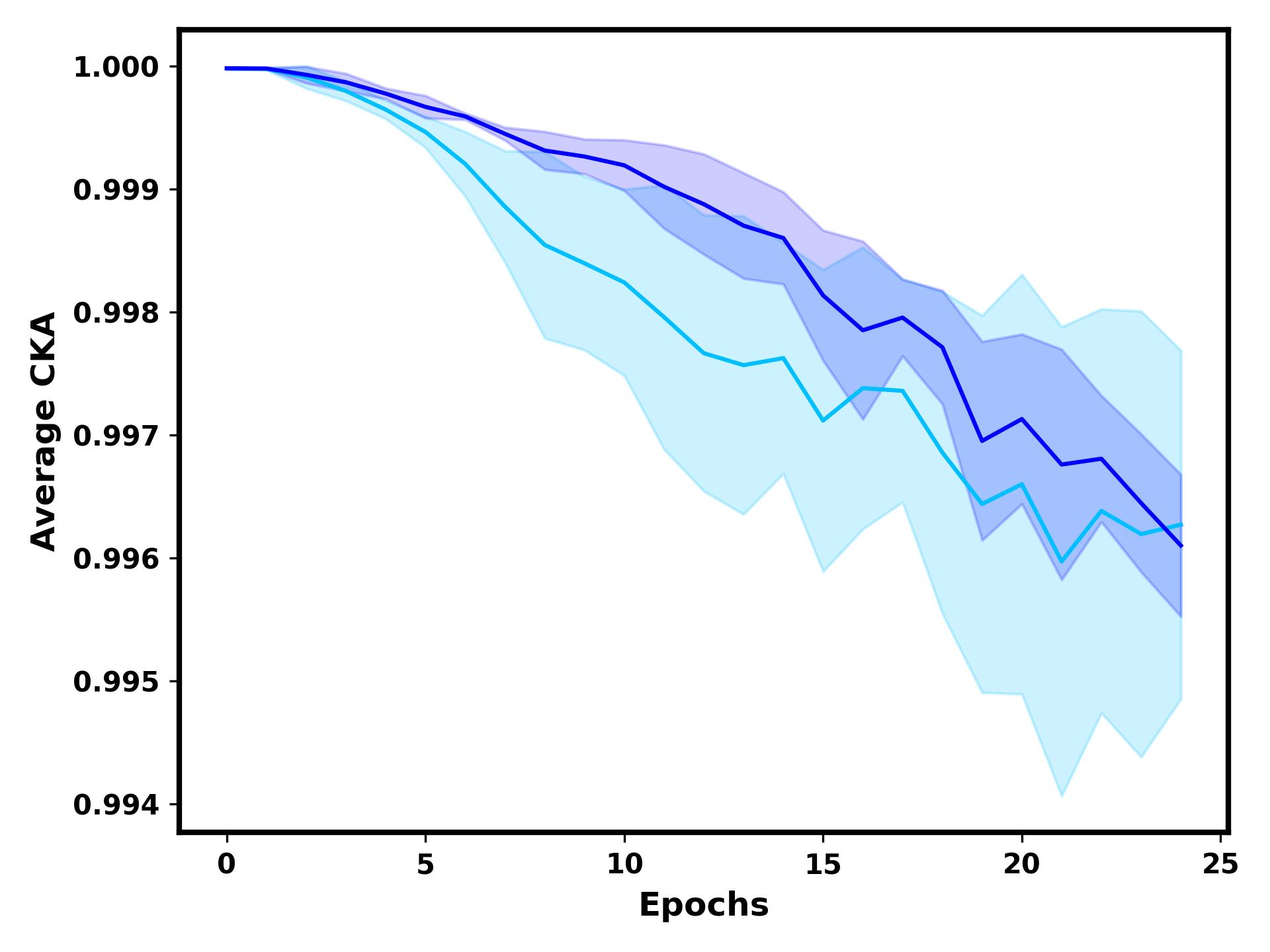}
        \caption{MLP (CIFAR-10 images)}
    \end{subfigure}
    \begin{subfigure}{0.32\textwidth}
        \centering
        \includegraphics[width=\textwidth]{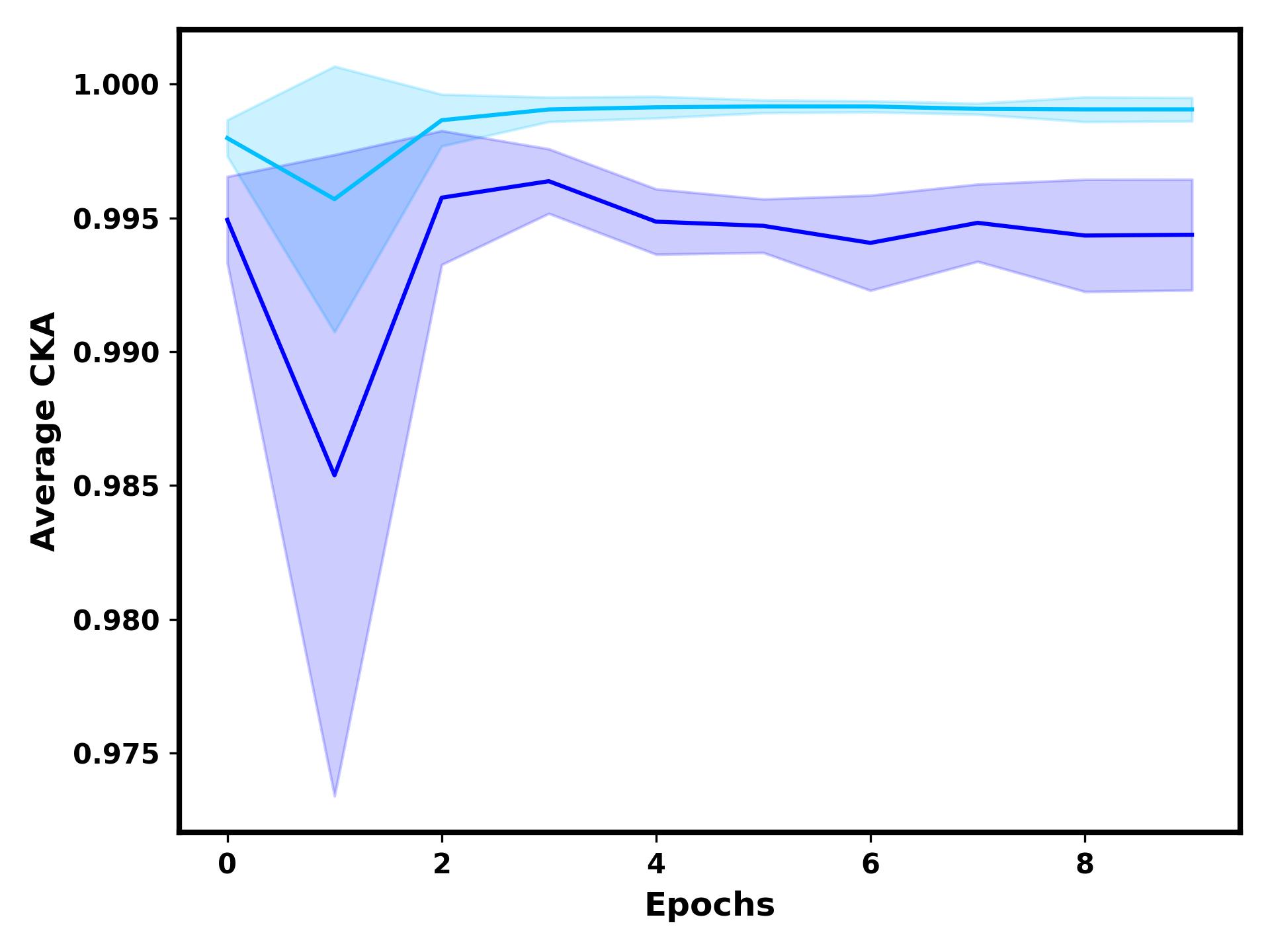}
        \caption{CNN (CIFAR-10 images)}
    \end{subfigure}

    \vspace{-0.5em}
    \caption{\textbf{Averaged CKA over five pairs of random GMNs, MLPs, and CNNs in Fashion-MNIST and CIFAR-10 classification.} 
    The \textcolor{skyblue}{skyblue} plot represents the averaged $\overline{\text{CKA}}(\textcolor{skyblue}{3},\textbf{a})$, while the \textcolor{blue}{blue} plot represents the averaged $\overline{\text{CKA}}(\textcolor{blue}{4},\textbf{a})$ for each architecture. Note that \textbf{the range of $\overline{\text{CKA}}$ varies significantly across datasets}.}
\end{figure}

\begin{figure}[h]
\centering
\includegraphics[width=0.2\textwidth]{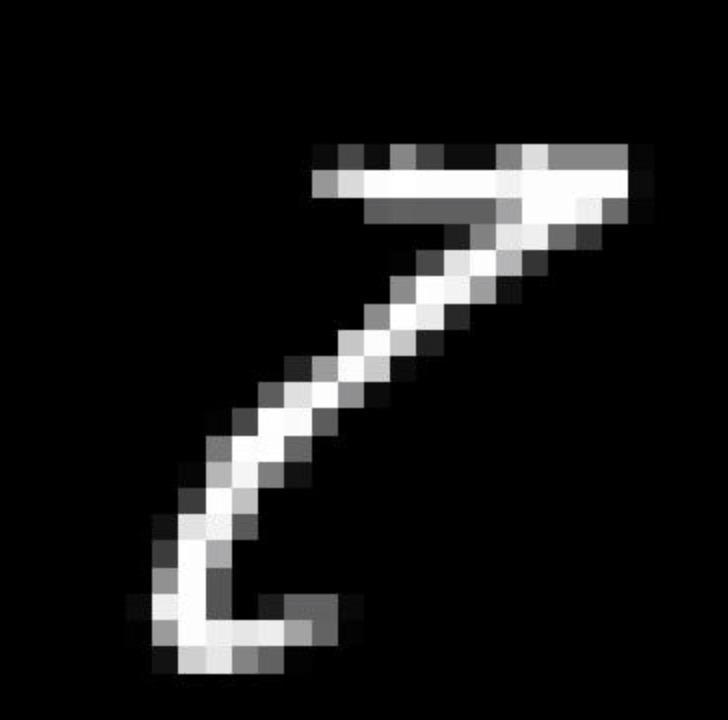}
\caption{\textbf{Sample analysis.} A sample in which the prediction of general NNs is always incorrect, while that of GMNs is always correct, across different initialization.}
\label{fig:incorrect}
\end{figure}

\end{document}